\documentclass{edm_article}
\usepackage{booktabs}
\usepackage{tcolorbox}
\usepackage{multirow}
\usepackage{graphicx}
\usepackage{enumitem} 
\usepackage{needspace} 
\usepackage{array} 
\begin{document}
\title{Objective Metrics for Evaluating Large Language Models Using External Data Sources
}


%
%
%
%

\numberofauthors{1} 
\author{
\alignauthor 
Haoze Du, Richard Li, Edward Gehringer\\
       \affaddr{Department of Computer Science}\\
       \affaddr{North Carolina State University}\\
       \affaddr{Raleigh, NC, USA}\\
       \email{hdu5, rli14, efg@ncsu.edu}
}

\maketitle

\begin{abstract}
Evaluating the performance of Large Language Models (LLMs) is a critical yet challenging task, particularly when aiming to avoid subjective assessments. This paper proposes a framework for leveraging subjective metrics derived from the class textual materials across different semesters to assess LLM outputs across various tasks. By utilizing well-defined benchmarks, factual datasets, and structured evaluation pipelines, the approach ensures consistent, reproducible, and bias-minimized measurements. The framework emphasizes automation and transparency in scoring, reducing reliance on human interpretation while ensuring alignment with real-world applications. This method addresses the limitations of subjective evaluation methods, providing a scalable solution for performance assessment in educational, scientific, and other high-stakes domains.
\end{abstract}


\section{Introduction}
With the advancement of Large Language Models (LLMs), their applications in fields such as text generation, automated evaluation, and educational feedback have become increasingly widespread \cite{chen2023language}). Particularly in higher education, LLMs are gradually being adopted to assist in grading, provide feedback, and enhance the student learning experience \cite{2010Improving}. However, how to effectively evaluate the performance of LLMs in these tasks remains an open question. Current evaluation methods mostly rely on benchmark tests, such as GLUE and SuperGLUE \cite{lyu2024evaluating}, or accuracy assessments based on human annotations. Yet, these methods may have certain limitations in educational contexts, such as failing to capture the reasonableness, relevance, and practicality of LLM feedback.

In the field of computer science education, graduate-level courses often involve complex design-oriented projects, such as Object-Oriented Design and Development (OODD). The learning objectives of such courses not only include mastering technical knowledge but also emphasize critical thinking, teamwork, and iterative development based on feedback \cite{2025Creating}. 
\vspace{1pt}
\hrule
\textit{This version of the paper is lightly revised from the EDM 2025 proceedings for the sake of clarity.}

In these courses, one workable and efficient way is to let students peer-review each other's project reports \cite{gehringer2007reusable}. This peer review mechanism can provide valuable feedback and help students understand evaluation criteria and improve project design \cite{2023Visualizing}. If LLMs can accurately simulate or enhance this review process, their potential for application in educational scenarios will significantly increase. Therefore, this study proposes a method that utilizes peer review data from an OODD graduate course as a benchmark to evaluate the review outcomes generated by different LLMs, aiming to explore the most suitable model for this task.

Existing research has explored the capabilities of LLMs in automated grading and feedback generation. For example, D. Bhatnagar \cite{bhatnagar2023fine} investigated the performance of GPT-3 in providing feedback on programming assignments and found that it was effective in detecting errors and offering general suggestions but lacked a deep understanding of task-specific contexts. Additionally, Reference\cite{chang2024survey} observed that feedback generated by LLMs often lacks consistency, potentially providing contradictory suggestions under different circumstances. Therefore, relying solely on traditional evaluation methods may not sufficiently measure the performance of LLMs in educational scenarios. 

Traditional evaluation methods for LLMs typically rely on standard datasets and automated metrics, such as BLEU, ROUGE, and BERTScore \cite{lyu2024evaluating}. While these methods offer certain advantages in assessing text generation quality, their effectiveness is limited when applied to complex tasks involving human interaction, such as educational feedback generation \cite{2024An}. For instance, evaluation methods based on automated metrics fail to comprehensively capture the reasonableness and practical impact of feedback \cite{bui2024chatgpt}. In educational settings, feedback not only needs to accurately identify issues but should also be constructive and provide specific suggestions for improvement \cite{fong2018feedback}.

The primary goal of this study is to propose and validate an LLM evaluation framework based on peer review data, specifically focusing on the following aspects:
\begin{enumerate}[topsep=0pt, partopsep=0pt, parsep=0pt, itemsep=5pt]
\item Constructing a benchmark dataset, as detailed below. 

\begin{table*}[h]
\centering
\caption{Some examples of tag data from peer reviews}
\label{tab:tag_examples}
\resizebox{\textwidth}{!}{%
\begin{tabular}{cccc}
\hline
Rubric prompt           & Comment from reviewer & Name of tag assigned    & Tag value \\ \hline
Are there any missing attributes for the admin? & No       & Contains explanation? & No        \\
Are there any missing attributes for a user?    & Credit card number was not asked for at any point & Contains explanation? & Yes \\
Are there any missing attributes for the admin? & No       & Positive Tone?        & No        \\
Are there any missing attributes for the admin? & Good job, I see them all.                         & Positive Tone?        & Yes \\ \hline
\end{tabular}%
}
\end{table*}

\item Fine-tuning LLM(s) to set up the metrics from the tagged data.

\item Evaluating the performance of the metrics from the fine-tuned LLM(s).
\end{enumerate}
The original data was collected from a graduate-level OODD course using Expertiza \cite{gehringer2007reusable} for multiple semesters and tagged by the students who participated in this class. Table \ref{tab:tag_examples} shows some examples of tags that were assigned.  The reviewer has filled out a review rubric, which consists of a set of \textit{prompts}.  For each prompt, the reviewer assigns a score and, optionally, provides a \textit{comment}.  The reviewee (author) is asked to assign appropriate tags to each review comment (such as, Contains explanation? or Positive tone?).  The \textit{tag value} is to be assigned appropriately by the author.

This framework aims to systematically assess the capabilities of LLMs in generating educational feedback and identify the most effective model for enhancing peer review processes in higher education.

The remainder of this paper is organized as follows. Section 2 reviews existing LLM evaluation methods and discusses relevant research on LLMs in educational feedback tasks. Section 3 introduces the methodology of this study, including dataset construction, definition of evaluation metrics, and experimental setup. Section 4 details the experimental design and presents a comparative analysis of feedback generated by different LLMs. Section 5 discusses the experimental results, analyzing the effectiveness and applicability of LLM-generated feedback. Section 6 addresses the limitations of the study and outlines potential directions for future research.  
Section 7 summarizes the findings and provides recommendations for optimizing LLMs in educational assessment.  

\section{Related Work}
\subsection{Overview of Evaluation Methods for Large Language Models}
The evaluation of Large Language Models (LLMs) primarily involves automated metrics, benchmark dataset assessments, and human evaluations. Traditional automated evaluation metrics, such as BLEU \cite{2002BLEU}, ROUGE \cite{lin2004rouge}, and METEOR \cite{banerjee2005meteor}, are widely used in natural language processing tasks. However, these metrics typically rely on surface-level similarity to reference texts and fail to adequately measure the logicality, coherence, and reasonableness of generated text \cite{chang2024survey}. In recent years, deep learning-based evaluation methods such as BERTScore \cite{zhang2019bertscore} and MoverScore \cite{zhao2019moverscore} have partially addressed this problem. However, they still face limitations when directly applied in educational scenarios \cite{gehrmann2022photon}.


Human evaluation remains an indispensable part of LLM assessment. For example, OpenAI adopted a method based on Reinforcement Learning from Human Feedback (RLHF)) \cite{liu2020personalized} in the development of GPT series models to improve the quality and acceptability of generated text. However, human evaluation often suffers from subjectivity, high costs, and difficulty in scaling up, prompting researchers to explore more objective and reproducible LLM evaluation methods.

\subsection{Rationality analysis of peer evaluation in design projects}
Peer assessment has been widely used across the curriculum for feedback and iterative improvement of student work. The benefits of peer evaluation include promoting students' critical thinking, increasing the diversity of feedback, and reducing the workload of instructors. In addition, research has shown that effective peer evaluation can help students better understand evaluation criteria and enhance their self-directed learning skills \cite{liu2006peer}.

In advanced courses, especially those involving complex project design (such as Object-Oriented Design and Development), peer evaluation is commonly used to assess the quality of project reports and provide  suggestions for improvement \cite{zeid2005peer}. However, research has found that students may have scoring biases, poor feedback skills, and inconsistent understanding of standards when conducting peer evaluations. Therefore, how to improve the rationality and effectiveness of peer evaluation has become an important research issue.

Recent research has explored automatic feedback generation based on LLMs to assist or replace manual peer evaluation. For example, Kulkarni et al. developed an automated evaluation system that combines machine learning and natural language processing techniques to provide targeted feedback to students \cite{kulkarni2024automated}.  Other studies focus on how to use LLMs to generate feedback that meets educational quality standards, such as providing specific recommendations, avoiding ambiguous evaluations, and using constructive language \cite{kasneci2023chatgpt}.

\subsection{The Application of LLM in Educational Evaluation}
LLM applications are becoming more widespread in the field of education, covering multiple aspects such as automatic grading, intelligent tutoring, and human-authorship detection \cite{zhai2022chatgpt}. For example, GPT-4, developed by OpenAI, has been used to generate feedback on academic writing and automatically score it \cite{kasneci2023chatgpt}. In addition, the Google research team proposed an intelligent education system based on PaLM 2, which can generate detailed feedback on student responses and predict possible misunderstandings \cite{anil2023palm}.

In peer-evaluation environments, LLM is mainly used to automatically generate feedback and assist teachers and students in improving the quality of evaluations \cite{gehrmann2023repairing}. Research has shown that using feedback generated by an LLM can significantly improve the objectivity of evaluations and reduce the subjectivity of assigning scores \cite{lu2024generative}. However, there are still consistency issues with the feedback generated by existing LLMs, such as repeatability issues, generating comments with different content or style when given the same input \cite{wu2024survey}.

This study introduces a framework for evaluating LLMs using peer-assessment data from masters-level design projects. By tagging key aspects of the feedback---such as rationality, operability, and consistency---the framework assesses the quality of responses generated by LLMs. This approach not only identifies the most suitable model for a given task but also lays a foundation for future research on LLM evaluation.

\subsection{DPO as a fine-tuning method}
Direct Preference Optimization (DPO) \cite{rafailov2024direct} is a model fine-tuning method that focuses on directly learning from user preferences rather than relying on indirect reward signals. Human preferences are assumed to follow the \textbf{Bradley-Terry model}, where the probability of preferring one response over another depends on the difference in their rewards:
\begin{equation}
  P(y_1 \succ y_2 | x) = \frac{\exp(R(x, y_1))}{\exp(R(x, y_1)) + \exp(R(x, y_2))}
\end{equation}
where \( R(x, y) \) is an unknown reward function, and \( y_1 \) is preferred over \( y_2 \).

By introducing a \textbf{KL divergence constraint}, the reward function is implicitly defined as the log-probability difference between the learned policy (\(\pi_\theta\)) and a reference policy (\(\pi_{\text{ref}}\)):
\begin{equation}
  R(x, y) = \beta \log \frac{\pi_\theta(y|x)}{\pi_{\text{ref}}(y|x)} + \text{const}
\end{equation}
where \(\beta\) is a hyperparameter controlling how much the learned policy can deviate from the reference model.

The problem of maximizing preference likelihood is reformulated as minimizing the following loss:
\begin{equation}
  \mathcal{L}_{\text{DPO}} = -\mathbb{E}_{(x, y_w, y_l)} \left[ 
    \log \sigma \left( 
      \beta \log \frac{\pi_\theta(y_w|x)}{\pi_{\text{ref}}(y_w|x)} 
      - \beta \log \frac{\pi_\theta(y_l|x)}{\pi_{\text{ref}}(y_l|x)} 
    \right) 
  \right]
\end{equation}
where \( y_w \) and \( y_l \) denote the preferred and dispreferred responses respectively, and \(\sigma\) is the sigmoid function.\\
By collecting user feedback through comparative choices, DPO adjusts the model's behavior to align better with user expectations. This approach is particularly effective in applications like recommendation systems and natural language processing, where understanding user satisfaction is crucial. DPO enhances user experience by optimizing model outputs based on explicit preference data, ultimately improving engagement and trust.

\section{method}
This study aims to fine-tune LLMs to evaluate the educational feedback provided by different LLMs using a peer-evaluation dataset of graduate design projects. The research methods mainly include dataset construction, definition of evaluation tags, and evaluation methods. This section briefly introduces the data processing flow, LLM evaluation methods, and quantitative tags used to analyze model performance.
\subsection{Research Flow Diagram}
\begin{figure}[htbp]
    \centering 
    \includegraphics[width=0.4\textwidth]{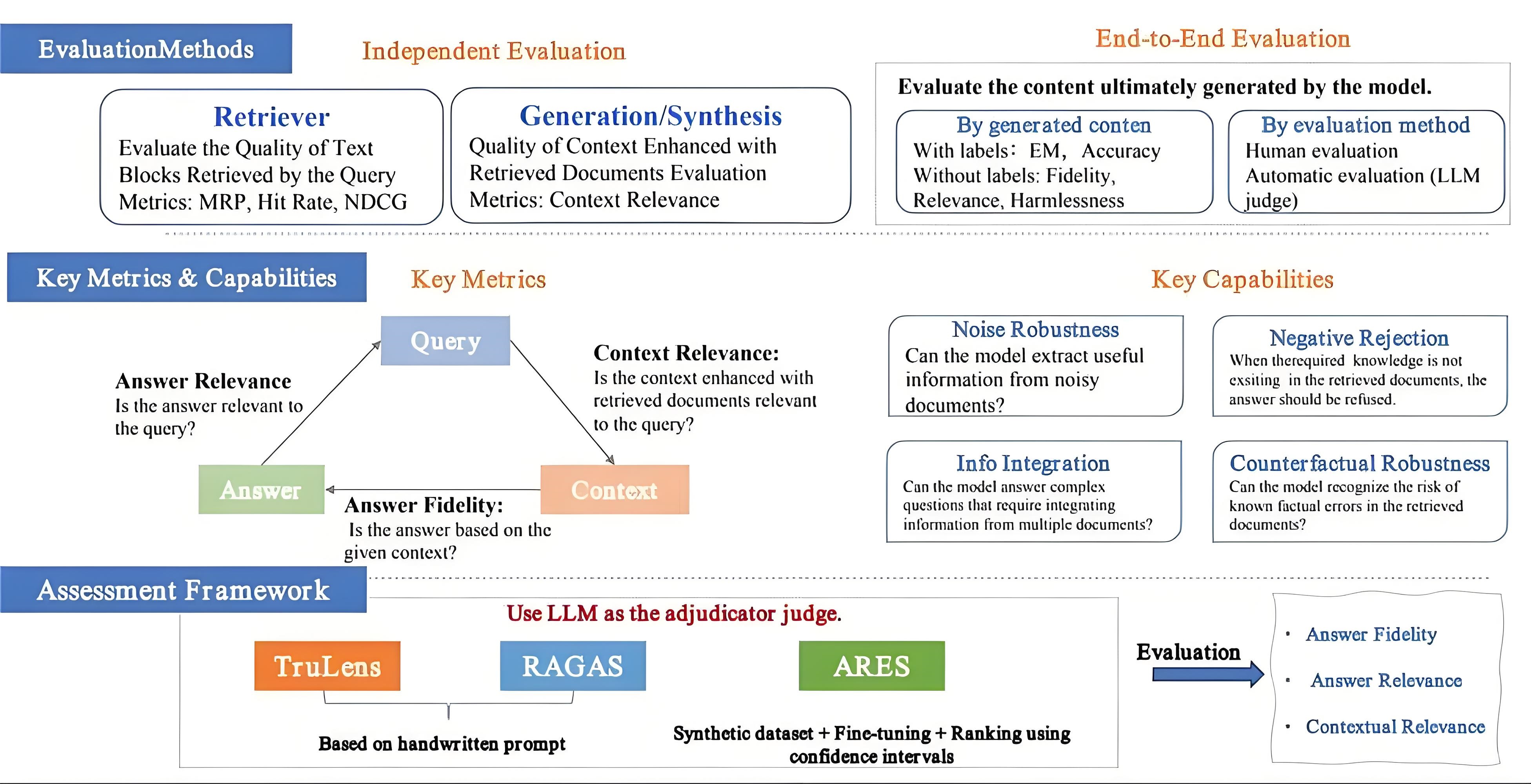}
    \caption{Research Flow Diagram} 
    \label{fig:Diagram} 
\end{figure}

A large-scale model is fine-tuned using a corpus of tags assigned by students in the object-oriented design course, to evaluate the output generated by another LLM. The process is shown in Figure \ref{fig:Diagram}. 

From the perspective of the evaluation method, in the independent evaluation, the retriever evaluation can measure the model's accuracy in retrieving information related to course projects based on these tags. For example, the ``Relevant" tag reflects the relevance between the retrieval and the project. The generation/synthesis evaluation focuses on the rationality and relevance of the evaluation content generated by the model based on these tags. In the end-to-end evaluation, when there are labels, the performance of tags like ``Suggests Actions" can be compared to the accuracy in the evaluation; for text that has not been labeled, tags such as "Uses Positive Tone" reflect the fidelity and relevance of the evaluation. Regarding the key tags, ``Answer Relevance" corresponds to ``Relevant", and ``Answer Fidelity" can be judged by tags like "Includes Explanation" and ``Consistent with Scoring". In terms of key capabilities, the performance of the model in tags such as ``Helpful" and ``Localized" can reflect its information integration ability and noise robustness. Finally, evaluation frameworks similar to TruLens, RAGAS, and ARES are used to comprehensively compare the performance of each model, thus judging how well the quality of the large-scale model has been evaluated. 

\subsection{Dataset Construction} \label{sec:dataset}
\subsubsection{Source of the data}
The dataset for this study is sourced from multiple semesters of a graduate-level course using the Expertiza peer-assessment system \cite{gehringer2007reusable}. The dataset consists of project reports submitted by students, accompanied by peer-review comments. These comments—also authored by students—offer evaluations and improvement suggestions, addressing various facets of the projects such as technological implementation, structural organization, and code quality. The raw data is retrieved anonymously, and includes: 
\begin{enumerate}[topsep=0pt, partopsep=0pt, parsep=0pt, itemsep=5pt]
    \item Project reports submitted by students (original PDF/text format).
    \item Items from the rubric to guide the students in peer-reviewing.
    \item Peer-review comments (structured text, including ratings and written feedback).
    \item Tags assigned by the reviewees to each comment, which name the tag and indicate its value (typically ``yes" or ``no", depending on, for example, whether a particular review comment contains an explanation).
    \item A ``credibility" metric, which attempts to measure how sure the researchers are that the tag has been given the correct value.
\end{enumerate}

\subsubsection{Data preprocessing}

Our dataset is derived from tags assigned by students in the Object-Oriented Design course.  After their work is peer-reviewed by their classmates, they are asked to assign labels (or ``tags'') to each of the comments in the reviews they have received.  Since we began tagging review comments in 2018, we have used 11 distinct tags in various semesters: Contains Praise, Identifies Problems, Offers Solutions, Uses Positive Tone, Mitigates Criticism, Localized, Helpful, Includes Explanation, Suggests Actions, Relevant, and Consistent with Scoring. The definitions of the 11 tags (``metrics'') are given in Table \ref{tab:evaluation_criteria}.

\newcolumntype{L}[1]{>{\raggedright\arraybackslash}p{#1}}

\newcolumntype{L}[1]{>{\raggedright\arraybackslash}p{#1}}

\begin{table}[h]  
    \centering 
    \caption{Evaluation Criteria}  
    \label{tab:evaluation_criteria}
    \begin{tabular}{p{0.7cm}L{2.5cm}L{4.5cm}} 
    \noalign{\vskip 3pt}
        \hline  
                \multicolumn{1}{p{0.7cm}}{\centering Tag\\\#} 
        & \multicolumn{1}{L{2.5cm}}{\centering\rule{0pt}{2.5ex}Name} 
        & \multicolumn{1}{L{4.5cm}}{\centering\rule{0pt}{2.5ex}Definition} \\\hline  
\noalign{\vskip 3pt}        M1               & Contains Praise               & Acknowledges strengths of the project.                  \\
        M2               & Identifies problems           & Points out shortcomings.                                  \\
        M3               & Offers solutions              & Provides suggestions for improvement.                    \\
        M4               & Uses positive tone            & Avoids negative language.                                 \\
        M5               & Mitigates criticism           & Lessens impact via tactful expression.                   \\
        M6               & Localized                     & Specific to the project.                                  \\
        M7               & Helpful                       & Substantial assistance for the reviewer.                  \\
        M8               & Includes explanation          & Explains reasons behind evaluation.                       \\
        M9               & Suggests actions              & Advises specific actions.                                 \\
        M10              & Relevant                      & Relates to project content.                               \\
        M11              & Consistent with scoring       & Aligns with provided score.                               \\
\noalign{\vskip 3pt}        \hline  
    \end{tabular}  
\end{table}

To ensure quality and consistency of the data, the following preprocessing steps were carried out in this study. First, A threshold $\left(\geq 0.35\right)$ for tag credibility \cite{akinepalli2024credibility} has been applied to keep the quality of the tagged data. Then, due to the limitations of fine-tuning different LLMs, a trade-off between cost and time was made by reducing the dataset to contain only 50 ``yes'' and 50 ``no'' tags for each metric.

To ensure the quality and consistency of the data, the following preprocessing steps were carried out in this study.

\begin{enumerate}[topsep=0pt, partopsep=0pt, parsep=0pt, itemsep=5pt]
\item \textit{Credibility threshold}
First, a threshold for ``tag credibility'' \cite{akinepalli2024credibility} ($\geq$ 0.3) was applied to maintain the quality of the tagged data. This step is crucial for filtering out unreliable or low-confidence tags, ensuring that the subsequent analysis and model training are based on accurate information.

\item \textit{Pattern removal}
Next, to further refine the dataset, only tagging data without discernible patterns (e.g., ``YNYN") were retained. This eliminates data that might be indicative of automated or less thoughtful tagging, thereby focusing on more nuanced and valuable human annotations.

\item \textit{Dataset reduction for LLM fine-tuning}
Finally, a trade-off between accuracy and time was made when reducing the dataset for fine-tuning different Large Language Models (LLMs). Recognizing the computational demands and the desire for timely results, the dataset was strategically scaled down. This involved randomly shuffling and selecting 50 ``yes" samples and 50 ``no" samples for each metric. This sampling approach was informed by current best practices for DPO with LLMs, where approximately 50 samples per tag are suggested as a sufficient size for effective fine-tuning, as indicated by OpenAI documentation. This balance ensures a manageable dataset size for efficient processing while still providing enough diversity and examples for the LLMs to learn effectively for specific tasks.
\end{enumerate}


\subsection{Dataset segmentation}
\label{sec:data}
The dataset was preprocessed and annotated with the 11 key tags from Table \ref{tab:evaluation_criteria}. To ensure the fairness of the evaluation, we divide the data in the following way:
\begin{enumerate}[topsep=0pt, partopsep=0pt, parsep=0pt, itemsep=5pt]
\item \textit{Training set} (60\%)  Used for fine-tuning some LLMs (such as Mistral-7B that supports LoRA training).
Provides high-quality peer-review samples for LLM to conduct supervised learning.
\item \textit{Validation set} (20\%)
As a tuning dataset for LLM feedback generation, it is used to adjust hyperparameters such as temperature, Top-$k$, and Top-$p$.
Ensures that the feedback generated by the model is semantically consistent with human feedback.
\item \textit{Test set} (20\%)
Mainly used for the final evaluation of feedback generated by the LLM.
\end{enumerate}
All data is shuffled so that data from different semesters are evenly distributed to the training, validation, and testing sets.

\subsection{LLMs Finetuning}
To make large language models (LLMs) create good evaluation metrics, the pre-trained models need to be tweaked so they can accurately judge different aspects of feedback quality. This process typically leverages domain-specific datasets, such as annotated peer review comments, to refine the model’s ability to classify and measure key evaluation criteria. Techniques such as Low-Rank Adaptation (LoRA) and supervised fine-tuning enable LLMs to learn from structured human evaluations, aligning their outputs with specific criteria like problem identification, suggested solutions, and tone analysis. Reinforcement learning from human feedback (RLHF) can further enhance the model’s scoring consistency by iteratively improving its ability to differentiate between constructive and unhelpful feedback. As a result, fine-tuned LLMs can produce reliable, standardized evaluation metrics that support automated assessments—minmizing bias while enhancing scalability and objectivity in educational and research contexts.

\section{Implementation}
This study experiments with some mainstream LLMs, including GPT-4o, Deepseek and Llama3. These models demonstrate exceptional performance in text generation and understanding tasks, making them suitable candidates for automatic review tasks and effectively meeting the requirements of various application scenarios.

\subsection{Setting up the Experiment}
To ensure the fairness of the experiments, we established a uniform prompt structure for the evaluating LLMs, as shown in the inset below.

After completing the fine-tuning process with DPO, the LLMs designated for evaluation are fully prepared and optimized to apply objective assessment metrics to outputs generated by various applications, including chatbots and other language models. Leveraging domain-specific datasets and reinforcement learning techniques, the models are capable of providing structured, quantifiable evaluations that minimize human bias while maintaining reliability. Their deployment enables automated, scalable evaluation of text-based AI outputs, facilitating improvements in generative model performance and refining responses based on well-defined quality standards. This approach enhances transparency and standardization in assessing AI-generated content across different applications and domains.

\begin{tcolorbox}[colback=gray!10!white, colframe=gray!80!black, title=\texttt{Prompt for fine-tuning LLMs}]
    \begin{verbatim}
prompt = (
f"Review_Comment: \"{review_comment}\"\n"
"Classify the following tags as json {tags}:\n"
"You should generate the tag value 
as \"{value}\", which -1 means negative
and 1 means positive.\n"
"Answer in JSON format as the {
    \"most_rel_tag\":
      {tag}, \"tag_value":{value}}."
)
    \end{verbatim}
\end{tcolorbox}

\subsection{Validating the Effectiveness of the Metrics}
Section \ref{sec:data} explains how the dataset has been segmented. We desire to compare three different methods of using the LLM:
\begin{itemize}[topsep=0pt, partopsep=0pt, parsep=0pt, itemsep=5pt]
    \item \textit{Direct} use of the LLM, without any extra training.  In this case, the LLM is simply asked to assess whether a review comment contains any of the 11 evaluation criteria from Table \ref{tab:evaluation_criteria}, e.g., does the comment identify a problem? Does it use positive tone?
    \item Use of the metric \textit{definitions} from from Table \ref{tab:evaluation_criteria}.  In this case, we use prompt engineering to tell the LLM what each of the metrics means, e.g., that if it says that a comment ``uses positive tone," then that comment should avoid negative language. 
    \item \textit{Fine-tuning} the LLM based on the tag dataset that was introduced in Section \ref{sec:dataset}. 
\end{itemize}
To compare these methods, we analyze key performance indicators such as agreement with human-labeled references, inter-method consistency, and robustness across different text inputs. Quantitative measures, including correlation scores and classification accuracy, along with qualitative insights from human evaluation, help determine which approach offers the most reliable and scalable assessment framework for evaluating AI-generated content.

\section{Results}
\subsection{Comparison Among Metrics with Test Set}
We have set up and fine-tuned multiple LLMs to conduct a comprehensive evaluation. GPT-4o is based on the online API using gpt-4o-2024-08-06 from OpenAI, providing access to its advanced capabilities in real-time, while DeepSeek (DeepSeek-r1-7b) and Llama 3 (Llama3-7b) are running locally, allowing for controlled experimentation and customization. Table \ref{tab:metric_comp} shows the accuracy of the test data.

\begin{table}[h]

\centering
\caption{Comparing accuracy of 3 methods on mainstream LLMs}
\label{tab:metric_comp}
\begin{tabular}{cccc}
\hline
\noalign{\vskip 3pt}\multirow{2}{*}{LLM} & \multicolumn{3}{c}{Methods}       \\ \cline{2-4} 
\noalign{\vskip 3pt}
                    & Direct & Definitions & Fine-tuning \\ \hline
\noalign{\vskip 3pt}
GPT-4o               & 75.24\%   & 75.34\%    & 79.82\%  \\
Deepseek             & 69.20\%   & 71.23\%    & 76.86\%  \\
Llama3               & 68.75\%   & 71.14\%    & 76.20\%  \\ \hline
\noalign{\vskip 3pt}
\end{tabular}%
\end{table}

From the results in Table \ref{tab:metric_comp}, it becomes evident that the fine-tuned approach outperforms both the direct-use approach and the metric definition method in accurately assessing generated text. Fine-tuned LLMs are specifically trained on annotated datasets, allowing them to develop a more nuanced understanding of evaluation criteria and consistently apply them across different text inputs. In contrast, the direct-use approach, where an unmodified LLM is prompted to assess quality, often produces inconsistent or overly generic evaluations, as it lacks targeted training for the specific task. Meanwhile, the metric definition method, which involves explicitly encoding evaluation criteria into prompts or system instructions, offers improved structure and alignment but still falls short in capturing contextual nuances and adapting to diverse text variations. By systematically comparing these methods, we observe that fine-tuned LLMs provide the most accurate assessments relevant to the given context, making them the superior choice for automated text evaluation tasks.

\subsection{Evaluating LLM-based Metrics with the Generated Contents}
To evaluate the effectiveness of the metrics derived from fine-tuned LLMs, we designed a structured task to assess the quality of feedback generated for given assignments. In this process, the fine-tuned models analyze student submissions and assign evaluation tags based on predefined criteria, such as clarity, constructiveness, and relevance. To ensure accuracy and reliability, a human instructor manually reviews the assigned tags, verifying whether they correctly reflect the feedback content. Any discrepancies between the model-generated tags and the instructor's judgment are recorded and analyzed to identify potential weaknesses in the LLM's evaluation process.

Table \ref{tab:example_real} shows some examples of the evaluation from both fine-tuned LLMs and human instructors in this class. LLM generates the contents, feeding the data described in Section \ref{sec:data}, and evaluated by human instructors and the fine-tuned LLMs.  In the table, 
\begin{itemize}[topsep=0pt, partopsep=0pt, parsep=0pt, itemsep=5pt]
\item the \textbf{first rubric item} is ``Has this team thoroughly tested at least one model and one controller?'' and the \textbf{comment} is, ``All the main models and views have been thoroughly tested. The naming convention for tests has also been properly followed. Good work!''\textbf{}
\item the \textbf{second rubric item} is ``Look at the newly-added code in the pull request. Check the variables methods and class names. List any name(s) that are not reasonable or suggestive of the functionality.'' and the \textbf{comment} is, ``rom the code that is visible in the Pull Request the code appears to be in good order and form.''
\item the \textbf{third rubric item} is ``Please review the code on Git. How well does the code follow good Ruby and Rails coding practices?'' and the \textbf{comment} is, ``It follows good coding practices to a great extent. Business logic is well separated from controllers. Just I feel that code from BorrowsController could have reduced in controller part and the applied business logic could have been shifted to corresponding model class and call model class methods by passing required params in controller.''
\end{itemize}













\begin{table}[h]
\centering
\caption{Examples of LLMs and students tagging the same comments}
\label{tab:example_real}
\resizebox{\columnwidth}{!}{%
\begin{tabular}{llccccc}
\hline
\noalign{\vskip 3pt}
\multirow{2}{*}{Examples} & \multirow{2}{*}{Tags} & \multicolumn{3}{c}{Fine-tuned LLMs} & \multicolumn{2}{c}{Human instructors} \\ \cline{3-7} 
\noalign{\vskip 3pt}
    &              & GPT-4o & DeepSeek & Llama3 & Instructor 1 & Instructor 2 \\ \hline
    \noalign{\vskip 3pt}
\#1 & Explanation? & Yes      & Yes        & No     & Yes            & Yes            \\
\#2 & Localized?   & No     & No       & No     & No           & No           \\
\#3 & Helpful?     & Yes      & Yes        & Yes      & Yes            & Yes            \\ \noalign{\vskip 3pt}\hline
\end{tabular}%
}
\end{table}

A total of 110 structured feedback examples, similar in format to those presented in Table \ref{tab:example_real}, were assessed to compare the accuracy of human evaluations and fine-tuned LLM-based metrics. These feedback samples were systematically categorized based on the 11 evaluation tags, with five positive and five negative examples for each tag, ensuring a balanced dataset for analysis. The evaluation process involved two parallel assessments: one conducted by human instructors, who manually reviewed each feedback instance and assigned appropriate tags, and another by the fine-tuned LLMs, which automatically classified the feedback according to the predefined criteria. The results from both methods were then compared to measure alignment, consistency, and potential discrepancies.  The table records how many times (out of 10 possible) the LLMs and the instructors agreed with the value of the tag in the dataset.  For example, GPT-4o agreed with the M1 tag in the dataset 9 times out of 10, whereas both human instructors agreed with it all 10 times.

\begin{table}[h]
\centering
\caption{Comparison of fine-tuned LLMs with human instructors. There are 10 cases for each tag (M1 to M11).}
\label{tab:final_eval}
\resizebox{\columnwidth}{!}{%
\begin{tabular}{cccccc}
\hline
\noalign{\vskip 3pt}
\multirow{2}{*}{Tags} & \multicolumn{3}{c}{Fine-tuned LLMs} & \multicolumn{2}{c}{Human instructors} \\ \noalign{\vskip 3pt}
\cline{2-6}
\noalign{\vskip 3pt}
                      & GPT-4o        & DeepSeek        & Llama3        & Instructor 1      & Instructor 2      \\ \hline\noalign{\vskip 3pt}
M1  & 9  & 8 & 6 & 10 & 10 \\
M2  & 9  & 9 & 8 & 10 & 10 \\
M3  & 8  & 8 & 8 & 9  & 9  \\
M4  & 7  & 8 & 8 & 10 & 10 \\
M5  & 8  & 8 & 6 & 10 & 9  \\
M6  & 7  & 6 & 6 & 10 & 9  \\
M7  & 9  & 6 & 6 & 10 & 9  \\
M8  & 7  & 6 & 6 & 10 & 10 \\
M9  & 10 & 7 & 7 & 10 & 10 \\
M10 & 10 & 7 & 8 & 10 & 10 \\
M11 & 7  & 8 & 6 & 9  & 9  \\ \noalign{\vskip 3pt}
\hline
\end{tabular}%
}
\end{table}

Table \ref{tab:final_eval} shows that fine-tuned LLM-based metrics can achieve performance levels comparable to human instructors in evaluating feedback quality. By systematically analyzing the alignment between model-generated assessments and instructor judgments, we observe a high degree of agreement across key evaluation criteria, such as problem identification, solution suggestion, and constructive tone. The fine-tuned models, trained on annotated peer review data, consistently apply predefined metrics, reducing subjectivity and variability often present in human evaluations. Additionally, statistical analysis shows that the model's tagging accuracy closely matches human-labeled benchmarks, with minimal discrepancies in cases requiring nuanced interpretation. These findings suggest that with proper fine-tuning and domain-specific adaptation, LLM-based evaluation metrics can serve as reliable, scalable alternatives to human assessment, offering efficiency and objectivity while maintaining human-level performance.

\section{Conclusions}
This study proposes an objective evaluation framework using peer-evaluation data from graduate design projects to measure how well different large language models (LLMs) perform in automatically generating educational feedback. We selected mainstream LLMs and conducted a comprehensive analysis of their generated feedback. We evaluated them based on 11 key tags, such as whether constructive suggestions were provided, whether a positive tone was used, and whether they were consistent with ratings. The experimental results show that the fine-tuned LLM-based metrics perform the best overall, outperforming other models in multiple dimensions such as feedback accuracy, relevance, localization, and rating consistency.

However, despite advancements in using LLMs for automated evaluation, several limitations remain. First, fine-tuned models may inherit biases from training data, leading to skewed assessments that reflect the subjective tendencies of human annotators rather than objective evaluation standards. Additionally, LLMs often struggle with interpretability, making it difficult to understand how they arrive at specific assessment scores, which reduces trust in their decision-making process. Scalability is another challenge, as fine-tuning requires large annotated datasets and significant computational resources, limiting accessibility for smaller research groups or institutions. Furthermore, LLM-based metrics may lack adaptability across domains, as a model fine-tuned for one type of evaluation (e.g., academic writing feedback) may not generalize well to another (e.g., creative writing assessment). Finally, there remains a gap between LLM-generated evaluations and human judgment, particularly in cases requiring deep contextual understanding, critical reasoning, or domain-specific expertise. Addressing these limitations will require improved fine-tuning methodologies, more transparent evaluation frameworks, and hybrid approaches that integrate LLM assessments with human oversight.


%
\bibliographystyle{abbrv}
\bibliography{sigproc}  
\balancecolumns
\end{document}